\documentclass[letterpaper, 10 pt, conference]{ieeeconf}

\usepackage{times}
\usepackage{graphicx}
\usepackage{array}
\usepackage{amsmath}
\usepackage{pseudocode}
\usepackage{booktabs}
\usepackage{multirow}
\usepackage{url}
\usepackage{balance}
\usepackage{bm}
\usepackage{units}
\usepackage{siunitx}
\usepackage{flushend} 

\usepackage[vlined, figure]{algorithm2e}
\SetAlgoSkip{medskip}

\usepackage{color}

\usepackage{xspace}

\begin{document}

\title{A New Vision for Smart Objects and the Internet of Things: \\ Mobile Robots and Long-Range UHF RFID Sensor Tags}

\author{\authorblockN{{\bf Jennifer~Wang} \\ \authorblockA{Massachusetts Institute of Technology}} \and
  \authorblockN{{\bf Erik~Schluntz} \\ \authorblockA{Harvard University}} \and
  \authorblockN{{\bf Brian~Otis} \\ \authorblockA{University of Washington}} \and
  \authorblockN{{\bf Travis~Deyle} \\ \authorblockA{Duke University}}
}

\maketitle

\begin{abstract}
We present a new vision for smart objects and the Internet of Things wherein mobile robots interact with wirelessly-powered, long-range, ultra-high frequency radio frequency identification (UHF RFID) tags outfitted with sensing capabilities. We explore the technology innovations driving this vision by examining recently-commercialized sensor tags that could be affixed-to or embedded-in objects or the environment to yield true embodied intelligence. Using a pair of autonomous mobile robots outfitted with UHF RFID readers, we explore several potential applications where mobile robots interact with sensor tags to perform tasks such as: soil moisture sensing, remote crop monitoring, infrastructure monitoring, water quality monitoring, and remote sensor deployment.

\end{abstract}

\section{ Introduction }

We can generally classify robot sensing into two modalities: remote contactless sensing (eg. lasers, cameras, and ultrasound) and direct touch (eg. haptics). Researchers have long speculated about a third sensing modality where ``smart objects'' or ``smart environments'' with embedded computation and sensing can directly measure and report salient information back to a robot. In more recent times, this general concept has garnered the moniker ``Internet of Things.''

UHF RFID is one compelling technology that speaks directly to the Internet of Things vision. Classic UHF RFID tags contain a small integrated circuit affixed to an antenna and mounted on a flexible substrate.  These battery-free tags harvest all of their power from the wireless signals that are also used for communication. The tags provide unique identification; are extremely low cost (sub-\$0.10 USD each); can be read from several meters away; can co-exist in the environment in the hundreds or thousands owing to low-level anti-collision protocols; and are produced in vast quantities each year for logistics applications.  Previously, researchers developed UHF RFID tags that also contain general-purpose computation as well as sensing capabilities \cite{buettner2008revisiting}. Similar sensorized UHF RFID tags are now commercially available. In this paper, we explore some early prototype applications for these tags and generally explore how these tags could be a boon for robotics.

The contributions of this paper are three-fold. First, we examine current commercially-available UHF RFID tags with sensing capabilities. We provide a basic comparison of this technology with other direct-sensing technologies such as Bluetooth Low Energy sensors, and we speculate about potential future robot applications.

\begin{figure}[t!]
  \centering
    \resizebox{0.95\columnwidth}{!}{
      \includegraphics[height=4in]{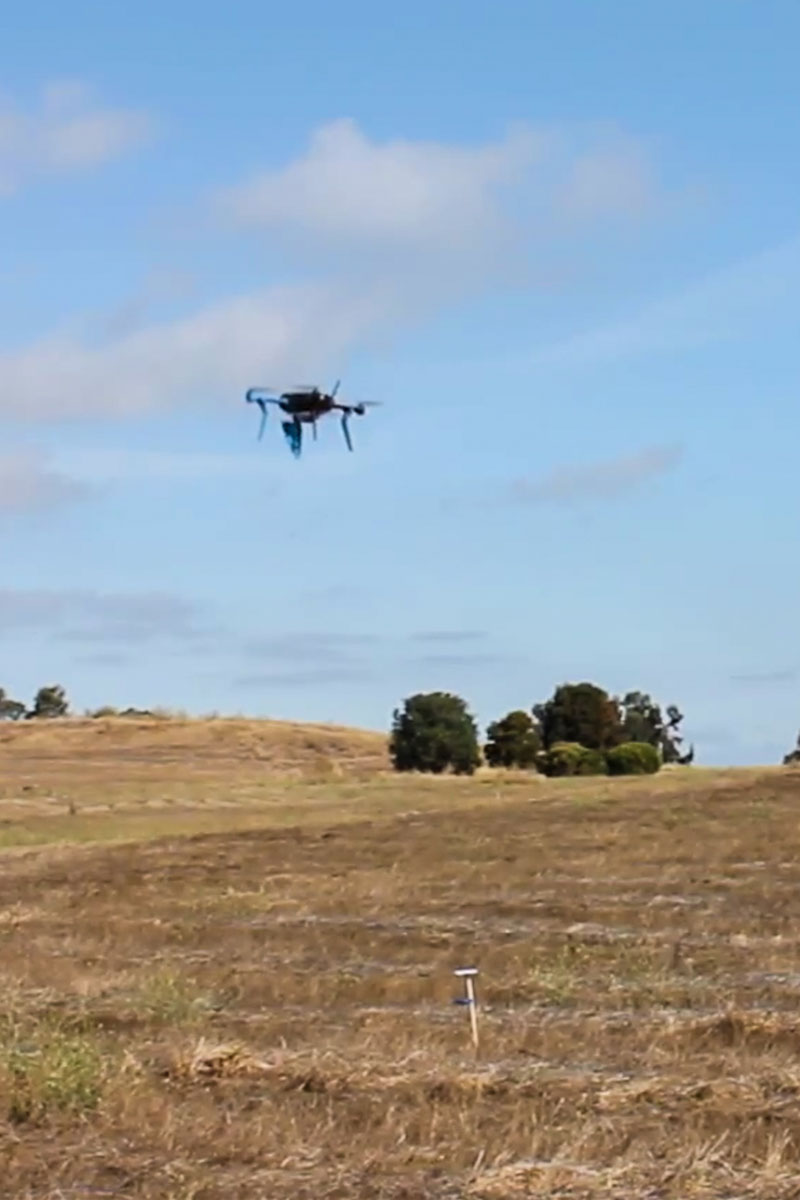}
      \includegraphics[height=4in]{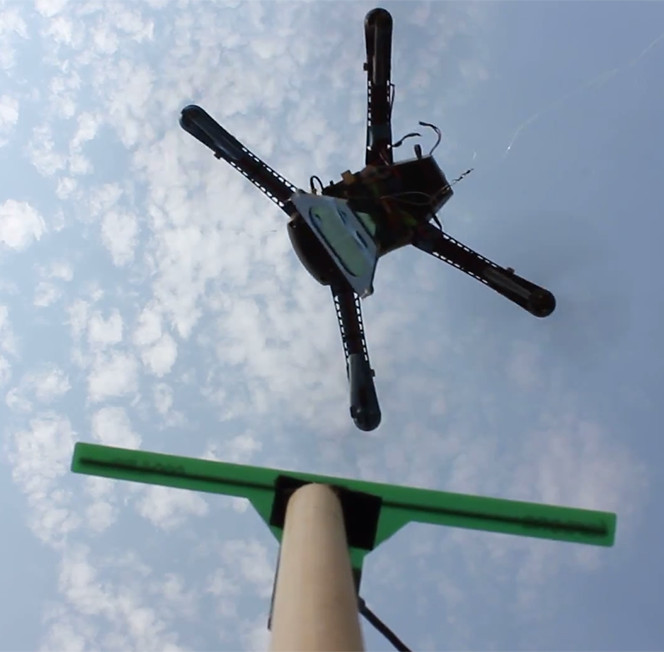}
    }
    \vskip 5pt
    \resizebox{0.96\columnwidth}{!}{
      \includegraphics[width=1.0\columnwidth]{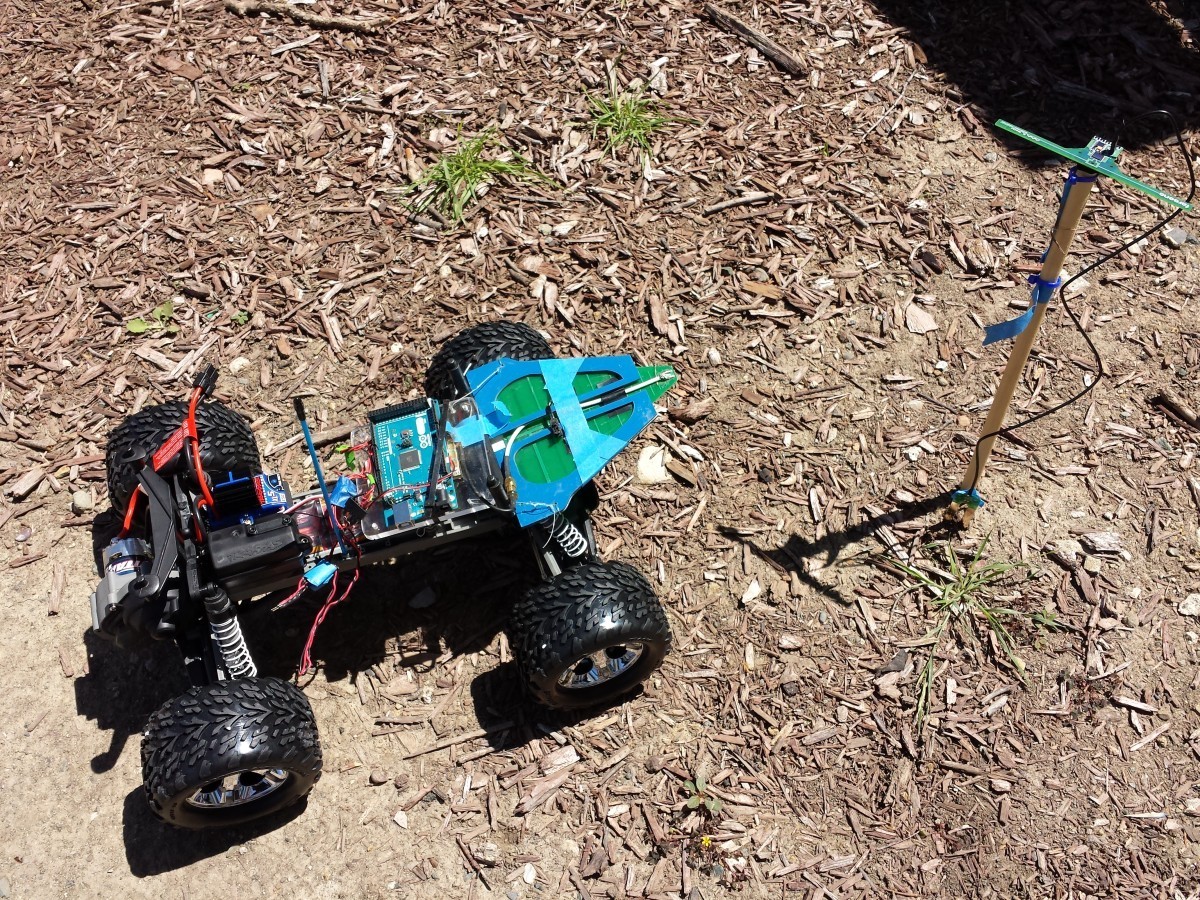}
    }
    \caption{\textbf{Top Row:} An unmanned aerial vehicle (a quadrotor drone)
      from 3D Robotics hovers above a Farsens Hydro tag mounted on a stick. The
      drone uses an attached UHF RFID reader to obtain direct soil moisture
      measurements from the tag. \textbf{Bottom:} An autonomous ground vehicle
      with a UHF RFID reader approaches a similar Farsens Hydro tag and obtains
      direct soil moisture measurements.}
    \label{fig:primary}
\end{figure}

Second, we demonstrate the first instance where robots interact with sensorized UHF RFID tags: a prototype application that uses commercial moisture-sensing tags for crop monitoring. We developed a pair of mobile robots (one aerial and one ground, as depicted in Figure \ref{fig:primary}), which we outfitted with UHF RFID readers for directly interacting with the sensor tags to obtain soil moisture measurements. We examine the requisite system components and discuss benefits of such a direct-measurement ``smart field'' compared to other remote sensing approaches to crop monitoring.

Finally, we report on a number of other experiments where we used an unmanned aerial vehicle (UAV) under RC control to (1) deploy sensorized UHF RFID tags into remote, out-of-reach locations; (2) perform a rudimentary type of infrastructure monitoring with sensorized tags affixed to a building's walls; and (3) use sensorized tags in a tree to perform rudimentary crop monitoring.

We recognize that many of these results are preliminary. We explicitly do not address the design of sensorized tags; perform a theoretical analysis of RF propagation; examine the design of the robots themselves; address RFID system design considerations such as antenna selection; nor do we provide a detailed evaluation of any one application (eg. soil monitoring) -- each of these considerations has already been covered in the literature \cite{buettner2008revisiting,finkenzeller_RFID_handbook,tdeyle_thesis}. Instead, our focus is to show, for the first time ever, how robots can use sensorized UHF RFID tags to facilitate a number of unique applications. Through these rudimentary prototypes, we hope to impress upon you, the reader, the potential benefits afforded to robots by these tags and provide a new tool for your toolbox. We believe there are myriad possibilities for sensor tags to be applied in other areas too, such as livestock monitoring, healthcare, and home automation. As the Internet of Things pervades our lives, it may even become practicable to embed tags in everyday objects, where: cups can inform the robot what liquid it contains and in what quantity; clothes can indicate their dirtiness; or food that can indicate its freshness or spoilage.  The rich information provided to robots by direct, embodied intelligence in the form of UHF RFID sensor tags could be transformative.

\begin{figure}[t!]
  \centering
  \includegraphics[width=0.95\columnwidth]{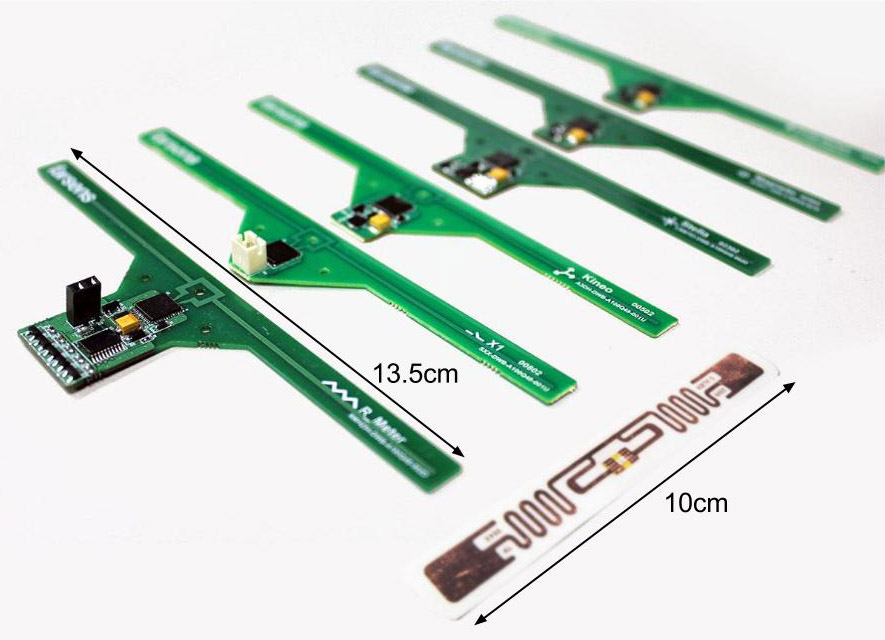}
  \caption{ A series of sensorized UHF RFID tags by Farsens (green). From left
    to right: resistance sensor, remotely-activated switch, 3-axis
    accelerometer, remotely-activated LED, magnetometer, pressure sensor. Also,
    a typical commercial UHF RFID tag by Alien Technologies (white) that provides
    unique identity only.}
  \label{fig:tag-types}
\end{figure}


\begin{figure}
  \centering
  \resizebox{0.95\columnwidth}{!}{
		\begin{tabular}{ r |  p{3.5cm} | p{3.5cm} }
	 	& Bluetooth Low Energy & UHF RFID \\ \hline
		Active Size & Postage Stamp & Grain of Sand \\ \hline
		\multirow{3}{*}{ Power } &	20mW during TX & no TX power \\
    & uW during sleep & uW at the tag \\
    & No “reader” power & 1W+ at reader \\ \hline
		Cost at Scale & \$3 ea. & sub-\$0.10 \\ \hline
		Lifetime & 4-10 years & indefinite (battery-free) \\ \hline
		Comms & Bidirectional \& push notifications & Bidirectional \\ \hline
    \end{tabular}
  }
\caption{ Comparing Bluetooth Low Energy and UHF RFID for direct sensing applications}
 \label{fig:rf-vs-ble-table}
\end{figure}

\section{ Related Work }

Roboticists have employed RFID to great effect. The unique identifier has proved useful for object recognition \cite{Gerkey08}, as a high-confidence landmark in SLAM implementations \cite{kleiner2006rtb}, for waypoint navigation \cite{bohn2004sdr}, and as a complementary sensing modality for multi-sensor fusion \cite{choi2009mobile}.  Many of these systems rely on low-frequency (LF) and high-frequency (HF) RFID tags, which have very short read ranges (5-10$\;$\SI{}{cm}). In contrast, ultra-high frequency (UHF) tags can be used for both short-range operation \cite{tdeyle_icra2013_inhand} as well as long-range operation out to several meters \cite{finkenzeller2003rh}. In robotics, UHF RFID tags have been used for robot localization \cite{joho2009modeling}, to locate tagged objects \cite{tdeyle_iros2014_search}, for medication delivery \cite{tdeyle_hri2013_medication}, and for manipulation \cite{tdeyle_iros2009_pps_tags}. 

Previously, researchers developed UHF RFID tags that contain general-purpose computation and sensing capabilities \cite{buettner2008revisiting}. Others have developed custom bio-monitoring tags \cite{tdeyle_rfid2013_ecg}, multimedia tags \cite{tdeyle_rfid2013_richmedia}, and moisture or temperature tags that can (for example) help detect inadvertent food thawing during transportation \cite{bhattacharyya2012low,siden2007remote,hamrita2005development}. These tags are also starting to see traction outside of academia; standards-compliant, sensorized UHF RFID tags are now becoming commercially available in small quantities from companies like Farsens and AMS. These tags have distinct advantages in terms of size, cost, and lifetime compared to other actively-transmitting battery-powered sensor technologies, such as sensors based on Bluetooth Low Energy; these advantages are summarized in Figure \ref{fig:rf-vs-ble-table}. For example, many sensing technologies are fundamentally limited by the capacity of on-board batteries, resulting in sensor lifetimes measured in weeks or months or systems that require burdensome periodic battery replacement \cite{beckwith2004report}. In contrast, battery-free UHF RFID tags can have lifetimes on the order of decades since they harvest nearly all of their operating power from a nearby RFID reader. By mounting RFID readers on robots and placing tags in the environment, system designers can leverage robot mobility to obtain the unique benefits afforded by UHF RFID sensor tags.

In the context of agricultural sensing, remote crop monitoring (usually via camera imaging) has already proved useful \cite{zarco2012fluorescence}, and it's even feasible to have UAVs obtain direct measurements for applications such as water quality monitoring \cite{thaler2014drones}. Inexpensive sensorized UHF RFID tags have the potential to augment or supplement existing remote-sensing systems to provide precise, direct measurements (eg. for calibration), and could even be used by existing farm equipment to opportunistically obtain relevant agricultural data.  Direct sensing can make agricultural systems safer, more efficient, more accurate, and more productive \cite{wang2006wireless,ruiz2009review}.  However, agricultural applications frequently involve large, expansive areas where wiring for communication and power is undesirable or impracticable. Wireless sensor networks have been used collect agriculture data by distributing sensors throughout a field, and transmitting information back to a base station \cite{ayday2009application}, sometimes using more-advanced mesh networking techniques \cite{beckwith2004report}. These sensor networks have been shown to provide actionable data that can improve growing conditions and irrigation schedules \cite{vellidis2008real}, but they still suffer from battery and cost constraints. UHF RFID sensor tags are a compelling approach to provide direct measurements that enhance other remote sensing techniques.

\section{System Components}
\label{sec:system}

For the purposes of this work, we focus on a prototype soil moisture sensing application.  Our system was comprised of three parts: the sensor tag, robot, and ground control software~(GCS).  In this section, we'll explore the relevant hardware and software components of all three.

\subsection{UHF RFID Sensor Tags}

\label{subsec:sensor-hardware}
\begin{figure}
  \centering
    \includegraphics[width=0.95\columnwidth]{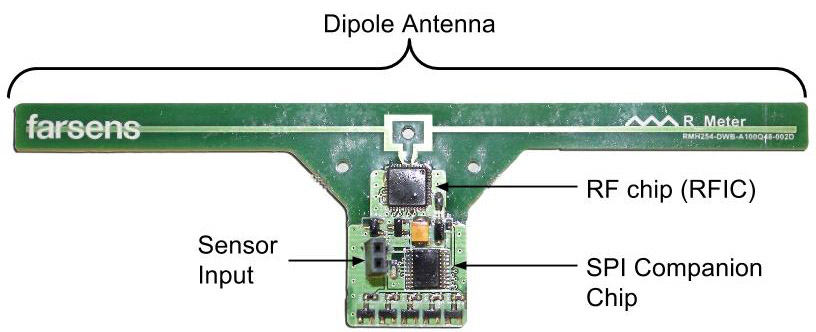}
    \vskip 5pt
    \includegraphics[width=0.95\columnwidth]{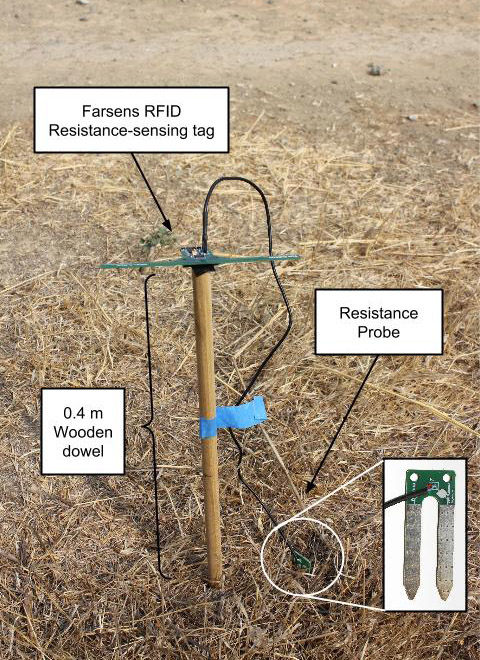}
    \caption{ \textbf{Top:} A commercially available sensorized RFID tag from
      Farsens. This tag contains a dipole antenna, RFID chip, SPI companion
      chip, and sensor input where resistance is measured. \textbf{Bottom:}
      Experimental setup of sensor tags. The tag was raised on a stake to
      increase read range. A resistance-measuring probe was inserted into ground
      to measure soil moisture. }
 \label{fig:farsens-photo}
\end{figure}

We employed sensor tags manufactured by Farsens, who produces a number of tags (Figure \ref{fig:tag-types}) for sensing: resistance, pressure, light, temperature, moisture, acceleration, magnetic fields, etc. For the purposes of our soil moisture sensing application, we used commercially-available Farsens Hydro Tags.  We also use Alien Omni-Squiggle tags as a control, though these tags only provide identity (ID) information.

Most UHF RFID tags work by harvesting RF power broadcast by a UHF RFID reader. Some tags (including the Farsens tags) have the option of also containing a rechargable battery and/or other power harvesting mechanisms for retaining state or data logging when RF power is unavailable; however, we used fully-passive (battery-free) tags for this work.  Our Farsens tags used a dipole antenna coupled to a specialized RF integrated circuit~(RFIC) created by Farsens that performs  power harvesting and communicates to-and-from the reader using Gen2 UHF RFID protocols. The RFIC provides electrical power (DC voltage) and a serial peripheral interface~(SPI) to a companion sensor chip.  In the case of the Hydro tag, the companion chip senses ambient temperature and electrical resistance between two leads of a probe inserted in the topsoil, thereby allowing for accurate soil moisture measurements -- as shown in Figure \ref{fig:farsens-photo}.  The tag also returns a 12-byte unique identifier (ID) in the same fashion as ID-only tags. To broadcast sensor measurements back to the RFID reader, the RFIC proxies the SPI data over the Gen2 UHF RFID protocol by reading from and writing to the tag's internal memory.

Other factors, such as the soil's chemical make-up and the ambient temperature, may also affect soil resistivity. A one-time calibration for the field of interest would probably be critical for practicable sensor deployments, but is outside the scope of this work. For the purposes of our work, we mounted Hydro tags to \SI{0.4}{m} high wooden stakes to improve signal reliability -- see Figure \ref{fig:farsens-photo}.

\subsection{Mobile Robot Hardware}

\subsubsection{Unmanned Aerial Vehicle (UAV)}

\begin{figure}
  \centering
    \includegraphics[width=0.95\columnwidth]{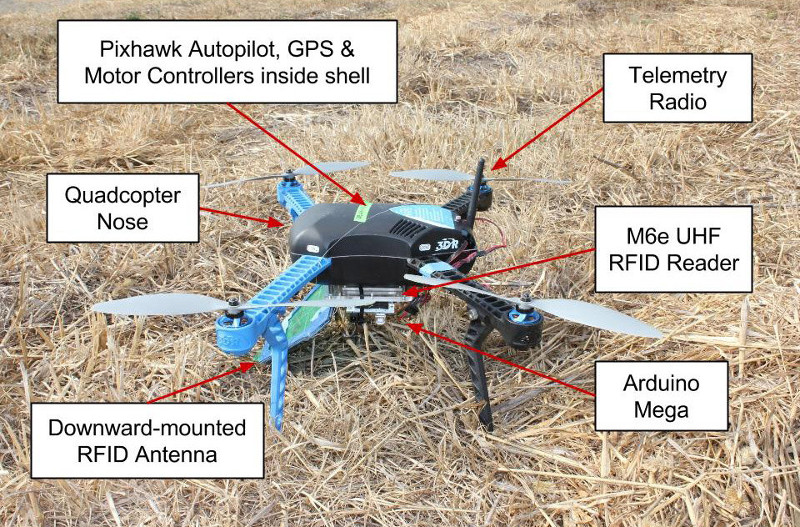}
    \vskip 5pt
    \includegraphics[width=0.95\columnwidth]{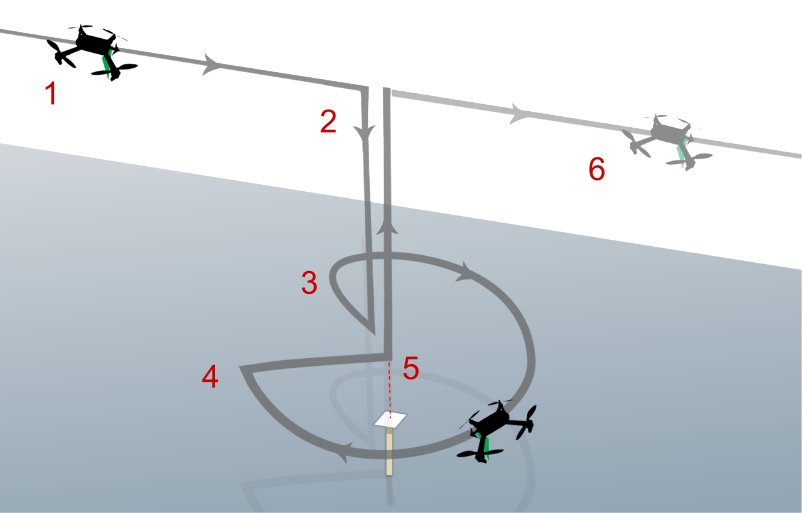}
    \caption{ \textbf{Top:} The UAV hardware system: RFID antenna, M6e UHF RFID
      reader, and Arduino Mega were mounted to a commercial IRIS drone from 3D
      Robotics. The IRIS drone includes a Pixhawk PX4 Autopilot, GPS and motor
      controllers. \textbf{Bottom:} Autonomous tag-reading behavior. While
      making repeated tag read attempts, the UAV (1) approached a GPS
      waypoint; (2) descended to an altitude of 1m; (3) hovered for 10 seconds; (4)
      flew a partial circle around the waypoint to search the local area; and (5)
      ascended back to \SI{3.5}{m} and proceeded to the next waypoint.  }
    \label{fig:uav}
\end{figure}

We employed a commercially-available 3D Robotics IRIS quadcopter as our unmanned aerial vehicle (UAV), as shown in Figure \ref{fig:uav}. During flight the UAV remained tethered at all times by a \SI{40}{lb} test nylon cable for both safety and regulatory compliance.  The drone remained in line of sight for all tests, and also featured an emergency human override via a separate \SI{2.4}{GHz} manual radio remote control (RC). A WA5VJB log-periodic antenna was mounted in a downward fashion to read tags as the quadrotor hovered nearby a tag.

\subsubsection{Unmanned Ground Vehicle (UGV)} 

\begin{figure}
  \centering
    \includegraphics[width=0.95\columnwidth]{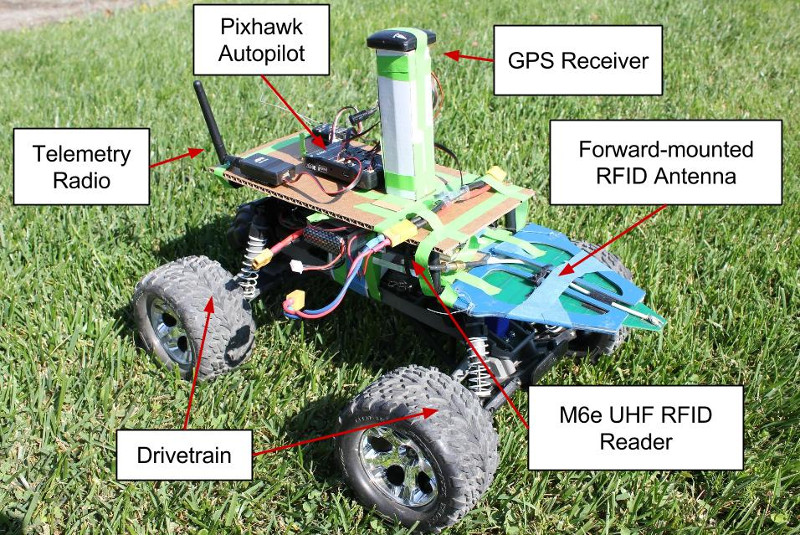}
    \vskip 5pt
    \includegraphics[width=0.95\columnwidth]{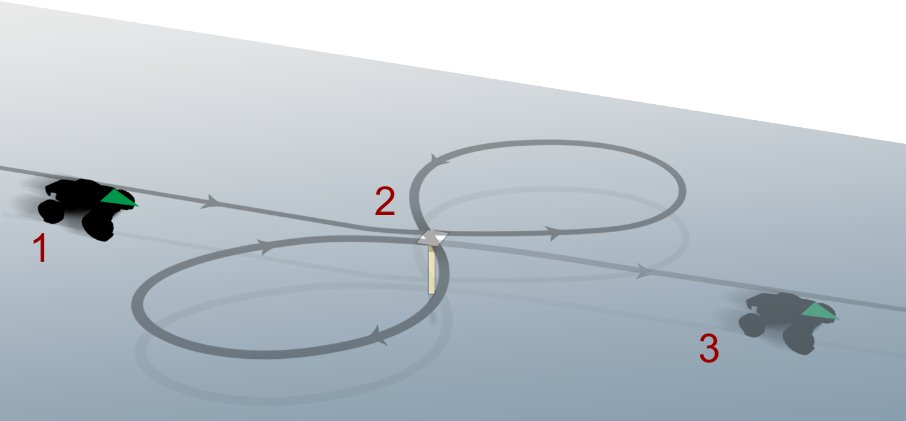}
    \caption{ \textbf{Top:} The UGV hardware system: RFID antenna, M6e UHF RFID
      reader, and Arduino Mega were mounted on a commercial RC car along with an
      added autopilot, radio, and GPS.  \textbf{Bottom:} Autonomous tag-reading
      behavior. While making repeated tag read attempts, the UGV (1) approached
      a GPS waypoint; (2) paused for 15 seconds and repeatedly circled around
      the waypoint for an additional 60 seconds; and (3) proceeded to the next
      GPS waypoint.}
    \label{fig:ugv}
\end{figure}

We employed a commercially-available Traxxas Stampede RC car as our unmanned ground vehicle (UGV), as shown in Figure \ref{fig:ugv}. The UGV included a sturdy chasis, drivetrain, suspension, rear wheel differential drive, electronic speed controller~(ESC), and a servo to control the steering of the front wheels. We removed the RC unit that came with the vehicle, and mounted a forward-facing WA5VJB log-periodic antenna to detect tags while approaching their location via ground.

\subsubsection{Hardware Common to Both Robots}

Both robots employed commercial Pixhawk PX4 autopilot systems with GPS telemetry as the on-board controllers. We used a ThinkPad T430 laptop as our ground control station~(GCS).

We mounted a commercial ThingMagic M6e UHF RFID reader with \SI{1}{W} RF output power, a WA5VJB log-periodic antenna, and an Arduino Mega to the underside of the UAV and to the topside of the UGV.  The RFID reader and antenna provide the core RFID functionality while the Arduino acts as an interface between the RFID reader and the Pixhawk autopilot. All three components (RFID reader, Arduino, and Pixhawk) communicate via UART serial interfaces; a coaxial cable connects the RFID reader with the log-periodic antenna.

\begin{figure}
  \centering
  \includegraphics[width=0.75\columnwidth]{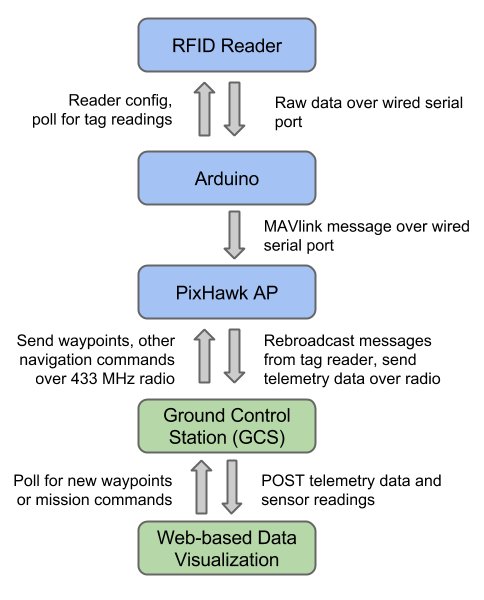}
  \caption{ Communication architecture of the system. Data from the RFID reader
    is passed through an Arduino, Pixhawk autopilot (AP), and the GCS back to the web
    dashboard. Commands from the web dashboard are passed through the GCS to the
    Pixhawk AP.  }
 \label{fig:comms-diagram}
\end{figure}

Figure \ref{fig:comms-diagram} shows the overall flow of data in the system. The Arduino interfaced directly to the UHF RFID reader and made repeated attempts to read nearby tags. The Arduino also interfaced to the Pixhawk autopilot running a slightly-modified version of the open-source ArduCopter (UAV) or ArduRover (UGV) firmware.  Upon successful tag detection, the Arduino checked the tag ID against a whitelist of known tags (a privacy precaution). If the tag was one of the known Hydro tags, the Arduino initiated a sensor read via the M6e and crafted a MAVProxy message (a standard message format) containing the tag ID and soil moisture measurement to the Pixhawk.  The modified Pixhawk firmware would forward this MAVProxy message along with standard telemetry MAVProxy messages (eg. GPS location) to our ground-station via a 433~MHz telemetry radio.

\subsection{ Ground Control and Visualization }

\begin{figure}[t!]
  \centering
    \includegraphics[width=0.95\columnwidth]{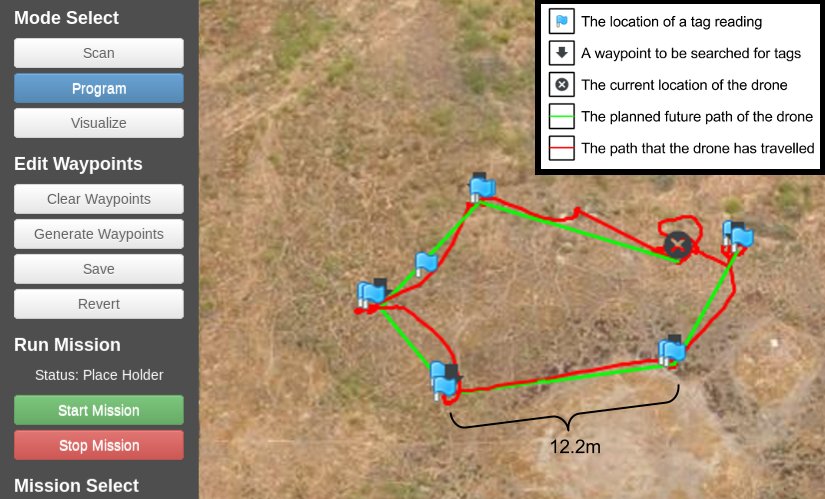}
    \caption{ Our custom Ground Control Station (GCS). The left hand menu gives
      user control over waypoints and other actions. Waypoints are marked with
      grey arrows and linked with a green line showing the trajectory of the
      drone. The past path of the drone is marked with a red line. Blue flags
      represent locations where a tag was read.  }
    \label{fig:gcs-example}
\end{figure}

We developed our own ground control station~(GCS) software that combined a slightly-modified version of the open-source MAVProxy ground station, a database, and a web-based GUI built atop public Google Maps APIs as shown in Figure \ref{fig:gcs-example}.  This ground control station served three purposes: 

\subsubsection{Mission Planning}

Using the web-based GUI, an end user could indicate a series of waypoints (candidate tag locations) or an area for the mobile robot to search (which would automatically generate waypoints).  This information was stored in the database as a ``mission.''

\subsubsection{Sending High-Level Mission Commands}

The modified version of MAVProxy obtained a mission (series of waypoints) from the database and generated a finite state machine~(FSM) to accomplish the mission.  The robot continuously transferred telemetry information to MAVProxy, including any sensor tag messages. Using the telemetry information and FSM, MAVProxy generated high-level commands for the robot; it transmitted the commands via the 433-MHz radio. Example high-level commands included actions such as: navigate to GPS coordinate, circle about a GPS coordinate, change altitude, takeoff, or land.

\subsubsection{Recording and Visualizing Telemetry}

MAVProxy continuously received telemetry messages from the robots (including GPS location messages and tag ID plus sensor measurement messages), which it stored in the database.  The web-based GUI updated its display based on new information.

\section{ Autonomous Robot Behaviors }

We developed a series of simple robot behaviors to read sensorized UHF RFID tags. These behaviors were based on a few high-level action primitives: navigate to GPS coordinate, circle about a GPS coordinate, change altitude, takeoff, or land. The combination of actions resulted in robot behaviors to detect tags and obtain sensor measurements.

\subsection{ Autonomous UAV Behavior }

The UAV search behavior is depicted in Figure \ref{fig:uav}. After taking off, the drone ascended to a an altitude of \SI{3.5}{m}. It then navigated to the first GPS waypoint (nominally corresponding to a tag location), traveling at \SI{150}{cm/s}. As the drone flew to a waypoint, it held pitch and roll neutral while adjusting yaw so that its nose pointed in the direction of travel.

For each waypoint, the quadcopter performed a search pattern to counteract GPS error (typically \SIrange{2}{3}{m}) and to improve RF connectivity to the tag. When the quadcopter detected that its GPS location was within \SI{1}{m} of the waypoint, it proceeded to hover. Then, it descended to 1.5~m relative to the ground (flat terrain) at a rate of \SI{25}{cm/s} while attempting to detect tags. The drone then hovered at \SI{1.5}{m} for \SI{15}{} seconds. If the tag was still not detected, the drone circled about the GPS coordinate with a radius of \SI{2}{m} at an altitude of \SI{1.5}{m}. After completing a \ang{270} circle without successfully finding a tag, the drone ascended to \SI{3.5}{m} at \SI{250}{cm/s}, exited the search behavior, and then either proceeded to the next  waypoint or exited the mission altogether. At any point, if the UAV received a positive tag reading, it immediately exited the search pattern and proceeded to the next waypoint.  

It also bears mentioning: The UAV based all altitude measurements off the initial takeoff location, and it used a barometer with an altitude precision of approximately $\pm$\SI{1}{m}. During the low-hover portions of the behavior, the drone could be within \SI{0.5}{m} of the ground based on the initial takeoff location. This poor altitude estimation placed severe constraints on our ability to autonomously read sensorized tags. This was a limitation of our chosen UAV, but could be mitigated in real deployments by better sensing (ultrasonic ranging, depth cameras, etc.) or better control algorithms. However, again, the drone's design is outside the scope of this paper.

\subsection{ Autonomous UGV Behavior }

The UGV search behavior is depicted in Figure \ref{fig:ugv}. The vehicle autonomously navigated to the first GPS waypoint (nominally corresponding to a tag location) such that it was facing the desired coordinates. The vehicle stopped when it was within \SI{1}{m} of the desired GPS location.  It waited for 15 seconds while attempting to obtain RFID reads.  If the tag was still not detected, the robot repeatedly performed a search pattern wherein it turned some random angle, moved away from the target GPS location, and then proceeded to return to (and face towards) the target GPS location. After approximately 1 minute of random retries, the UGV exited the search behavior and either selected another waypoint or exited the mission altogether. At any point, if the UGV received a positive tag reading, it immediately exited the search pattern and proceed to the next waypoint.

\section{ Experiments and Results }

\begin{figure}
  \centering
  \includegraphics[width=0.95\columnwidth]{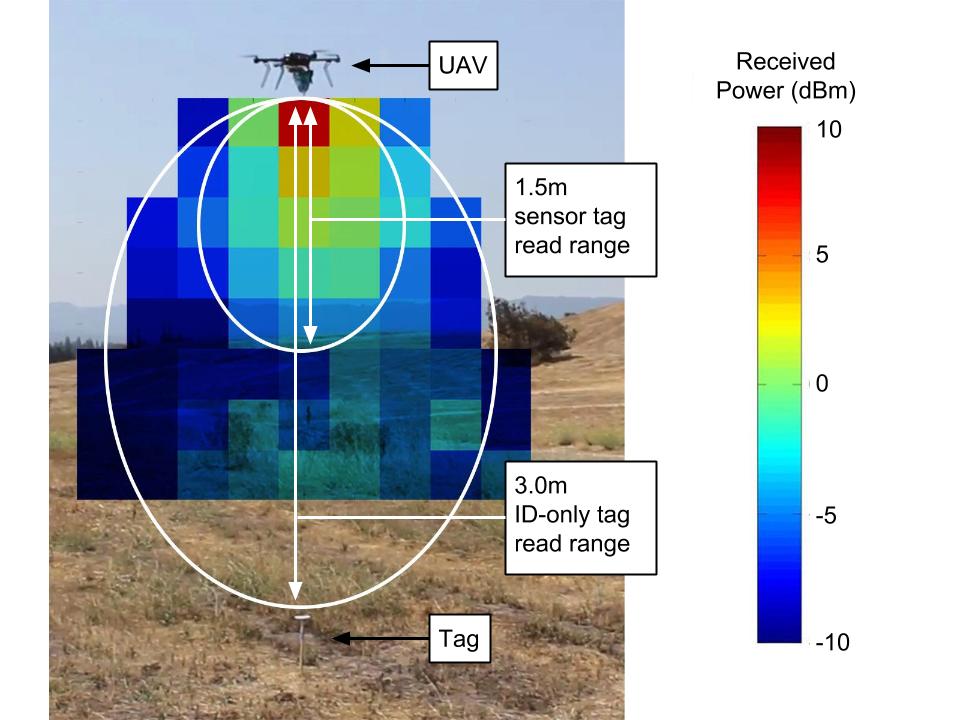}
  \caption{Measured RF power at the tag RFIC terminals gives an estimate of read
    range. In the lab, we attached a WA5VJB log-periodic antenna transmit
    antenna to a \SI{1}{W} RF source and measured the received power obtained
    using a Laird UHF S9025-PL patch antenna (circularly polarized, \SI{5.5}{dB}
    gain). The Farsens tags' dipole antenna (linearly polarized) has 2dB less
    gain, so the RF power obtained by real tags would be at least 2dB lower
    compared to the obtained measurements. In lab, we observed successful RFID
    sensor tag measurements when the RFIC received in excess of \SI{-5}{dBm}; owing to
    antenna differences, we expect positive dipole tag readings at distances
    where the patch antenna received in excess of \SI{-3}{dBm}. This corresponds to a
    \SI{1.5}{m} read range for the dipole Farsens tags, which matches what we
    saw in our experiments. The ID-only tags required less power and we were
    able to achieve a read range in excess of \SI{3}{m}. A more complete
    accounting of RF budgets and antenna selection considerations can be found
    in \cite{tdeyle_thesis}.}
  \label{fig:power}
\end{figure}

We performed a series of experiments with both robot platforms using sensorized tags as well as ID-only tags.

\subsection {UAV Experiments}

\begin{figure*}[t]
  \centering
  \resizebox{1.95\columnwidth}{!}{
    \includegraphics[height=3in]{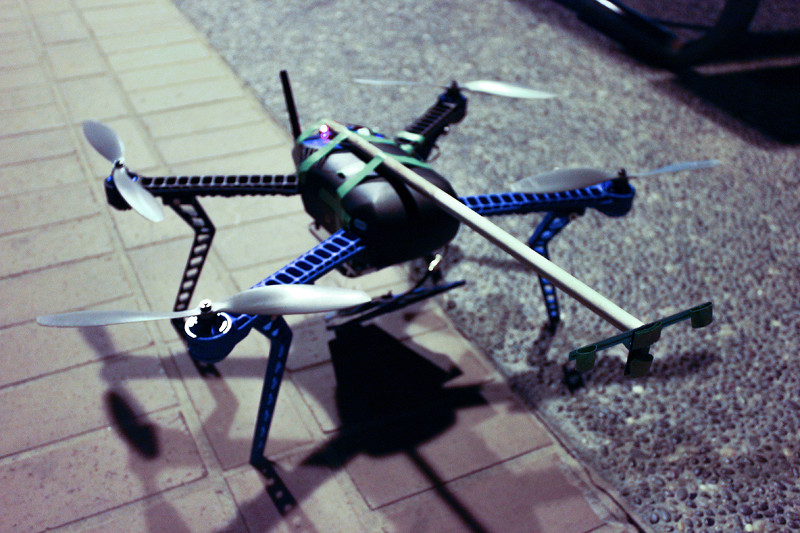}\hspace{2pt}
    \includegraphics[height=3in]{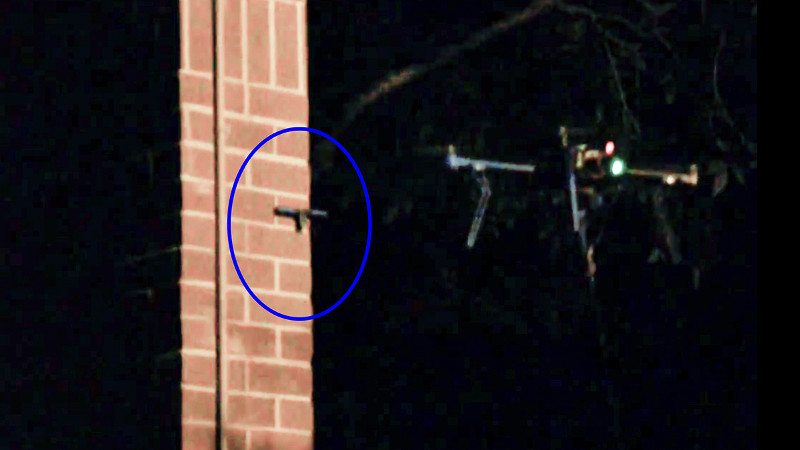}\hspace{2pt}
    \includegraphics[height=3in]{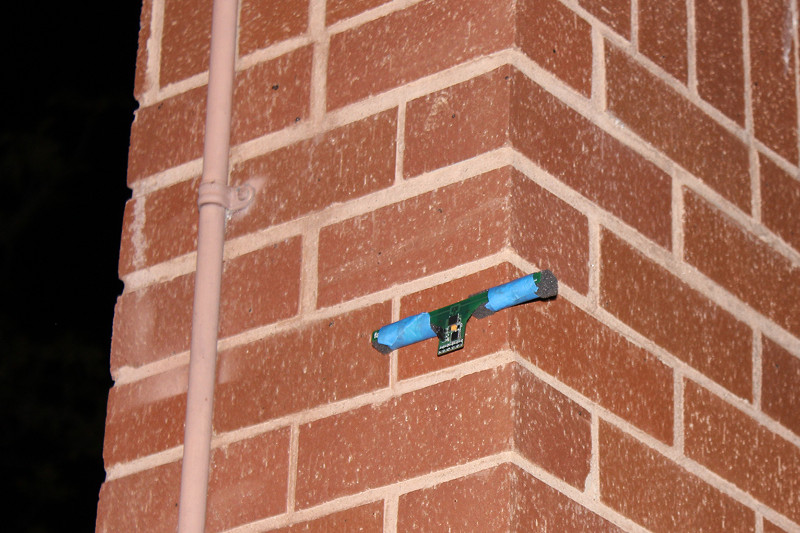}
  }
  \caption{ Mobile robots could be used to affix sensors in hard-to-reach or
    difficult-to-service locations, such as retrofitting buildings, bridges, or
    treetop canopies with sensor tags. In this case, we outfitted a UAV with a
    boom arm tipped with an adhesive-backed sensor tag (left). Under RC control,
    the UAV deployed the tag and could return later to take measurements (middle
    and right).}
  \label{fig:wall-placement}
\end{figure*}

\subsubsection{Autonomously Reading Sensor Tags}

Early lab experiments with the log-periodic transmit antenna (like the one carried by the drone) and a patch receive antenna (not used) suggested that the sensorized Farsens tags would have a best-case read range of just \SI{1.3}{m} (Figure \ref{fig:power}) -- and perhaps less depending on antenna polarization mismatch. Coupled with GPS position errors and barometric altitude estimation errors, we knew that autonomous reading of sensor tags would be challenging. Because of these constraints, we were unable to reliably read the sensor tags from the UAV under autonomous control, though we were able to read them under remote control as discussed later. Improvements to the drone's sensing or state estimation, and alternative antenna configurations on the drone or tag may have permitted autonomous sensor reading behaviors; however, these detailed design considerations are well-known in the literature \cite{tdeyle_thesis} and were outside the scope of this paper.

\subsubsection{Autonomous Reading of ID-Only Tags}

We planted five Alien Omni Squiggle tags in a $40\times40$\SI{}{m} open grass field at least \SI{10}{m} apart.  The flying area was free of obstructions taller than \SI{1.5}{ft}, and wind was less than \SI{4}{mph}. We pre-recorded each tag's GPS location into our ground control station and executed the autonomous UAV behavior from the previous section. We ran two trial missions, which resulted in positively detecting 5/5 tags (trial 1) and 4/5 tags (trial 2) using the ID-only tags.

\subsubsection {Remote Control Reading of Sensor Tags}

We planted three Farsens Hydro tags in the same $40\times40$\SI{}{m} open grass field at least \SI{10}{m} apart. We manually controlled the quadrotor using a \SI{2.4}{GHz} radio.  We manually flew between the tags at an altitude of \SI{3.5}{m}, descended to \SI{0.5}{m}, and hovered over each sensor tag until we obtained a sensor measurement. Obtaining a sensor measurement from the tag during hover could take up to 30 seconds owing to drone stability under manual control, antenna orientation, and tag charge-up time. In the end, we were successful at reading all 3/3 sensor tags under RC control.

\subsection {UGV Experiments}

\subsubsection{Autonomously Reading Sensor Tags}

The UGV had fewer control issues compared to the UAV.  We planted three Farsens Hydro tags in the field at least \SI{10}{m} apart and gave their GPS coordinates as waypoints to the ground control station. The tags were elevated on stakes so that the tag would be at the same height as the RFID antenna mounted on the vehicle.  Because of the shorter read range of the Farsens tags and GPS error, we were only able to read 2/3 of the tags during our experiment.

\subsubsection{Autonomously Reading ID-Only Tags}

We also tested the UGV platform's ability to autonomously read ID-only commercial tags with a longer read range. We placed three ID-only tags in a field approximately \SI{10}{m} apart and gave their GPS coordinates as waypoints. Because of their longer read range, we placed them directly on the ground rather than elevating them with stakes. The UGV successfully navigated to and read all 3/3 ID-only tags.

\subsubsection{Remote Control Reading of Sensor Tags}

We placed three sensor tags in a field approximately \SI{10}{m} apart. The tags were elevated on stakes so that the tag would be at the same height as the RFID antenna mounted on the car. We then manually drove the UGV to each of tag locations and obtained a reading from each tag. We successfully obtained sensor measurements from all 3/3 tags.

\section{ Additional Exploratory Experiments }

In addition to soil moisture monitoring, we explored a series of ``proof of concept'' applications that can help illustrate the diverse application areas for outdoor sensing using UHF RFID tags.

\begin{figure}
  \centering
  \resizebox{0.95\columnwidth}{!}{
    \includegraphics[height=2.4in]{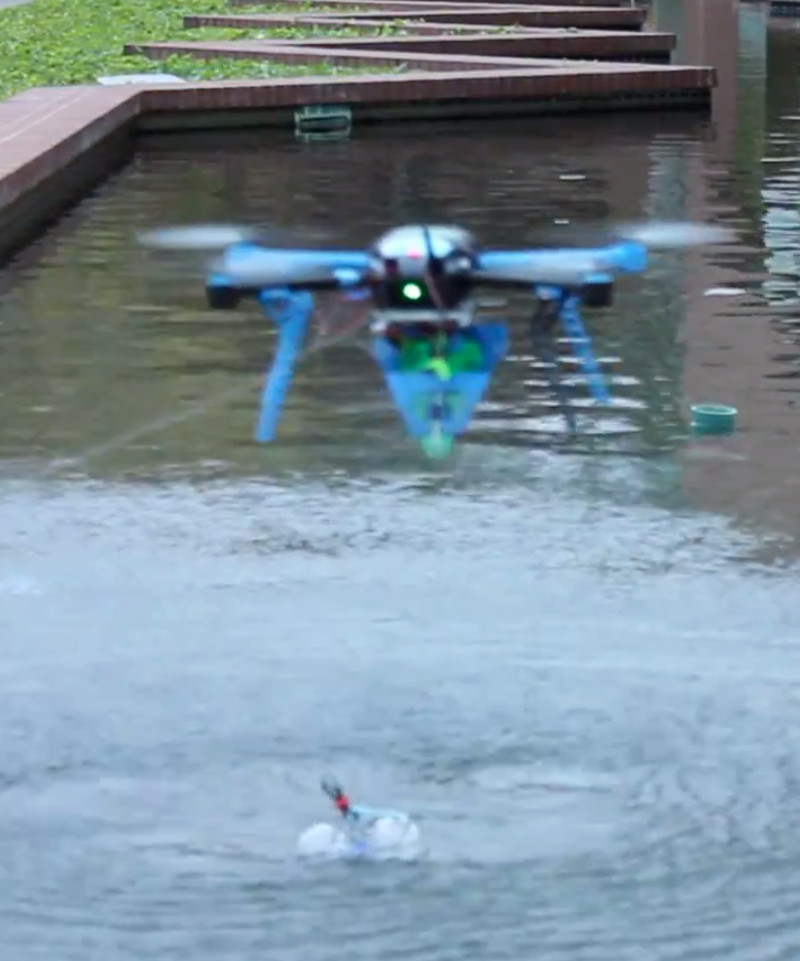}\hspace{2pt}
    \includegraphics[height=2.4in]{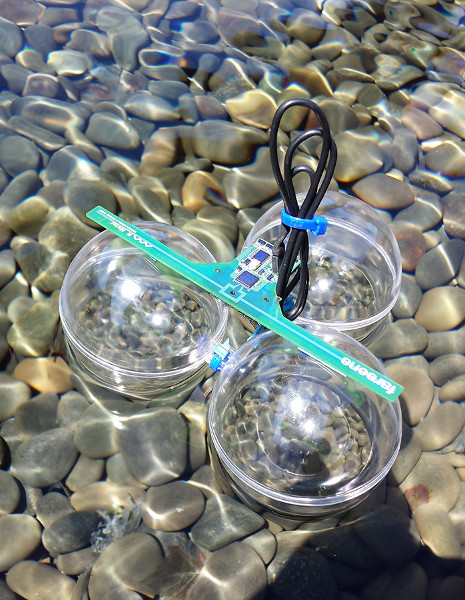}
  }
  \caption{ Drone reading a sensorized tag to assess water quality. Drones could
    take sensor measurements in difficult to reach locations over water.  }
  \label{fig:water}
\end{figure}

\subsection{ Deploying Sensor Tags }

One compelling use case for battery-free sensor tags is to place tags in hard-to-reach locations where direct human measurements or battery replacement would be difficult (eg. on building exteriors, in treetop canopies, etc.). As shown in Figure \ref{fig:wall-placement}, we attached a boom arm to our UAV, tipped with an adhesive-backed sensor tag.  Under RC control we flew the quadrotor up to a wall and ``poked'' the wall with the tag-tipped boom arm.  The adhesive on the tag pulled it off the boom arm and left it placed in the hard-to-reach location; we confirmed our ability to read the sensor tag by re-flying to the location with the drone-mounted reader.

\subsection{ Water Quality Monitoring }

As shown in Figure \ref{fig:water}, we attached one of the resistance-measuring tags to a flotation device and took measurements with an RC-controlled drone.  We suspended the tag's probe into the water and obtained conductivity measurements (a proxy for salinity). While this initial implementation is quite rudimentary (and the sensor ill-tuned to the application), the notion of using battery-free sensor tags with virtually-infinite lifetimes has some interesting implications for robots performing long-duration water quality monitoring -- where tags could either be floated en masse or attached to common anchored locations.

\subsection{ Infrastructure Monitoring }

\begin{figure}[t!]
  \centering
  \includegraphics[width=0.95\columnwidth]{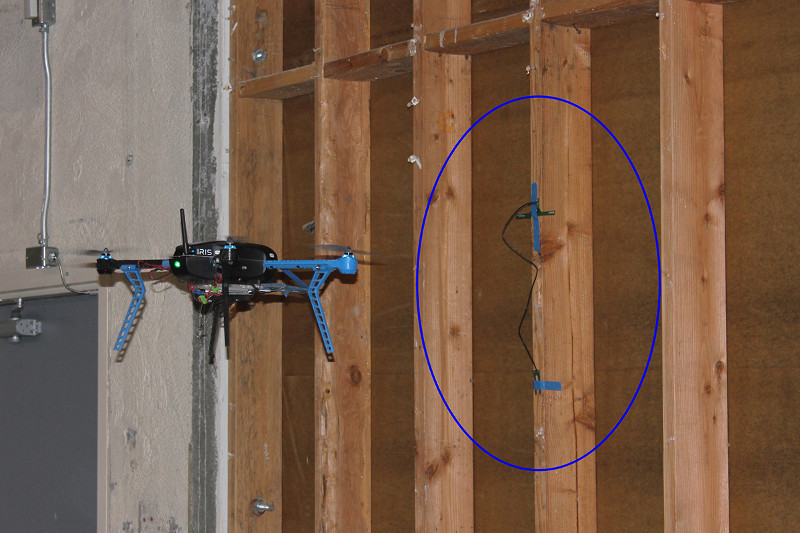}
  \caption{ Drone taking a measurement for infrastructure monitoring. Sensorized
    tags could be placed in the support structures of buildings, the trusses
    under bridges, or other critical yet hard to reach locations.  }
\label{fig:indoor-read}
\end{figure}

One oft-discussed applications for drones is to perform infrastructure monitoring: measuring stress, strain, corrosion, wear, etc. for hard-to-reach locations on buildings, bridges, power lines, and dams. Non-contact remote sensing using cameras and lasers is certainly a compelling capability. But direct-measurement using sensor tags offer the possibility to take direct sensor measurements (eg. for calibration), or even to obtain measurements inside the structures by using tags embedded inside during their construction.  As a rudimentary example, Figure \ref{fig:indoor-read} shows a drone under RC control taking measurements from a basic tag on the wall of a building.  In this case, the drone measured the moisture content in a building support beam, but it would have been equally viable to measure other useful information such as strain or stress.

\subsection{ Crop Monitoring }

\begin{figure}[t!]
  \centering
  \includegraphics[width=0.75\columnwidth]{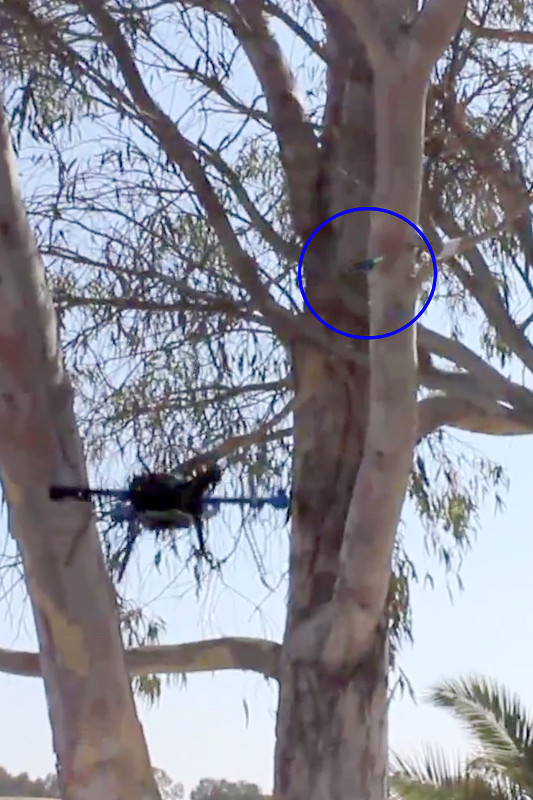}
  \caption{ Drone reading a sensorized tag in a tree. Drones could be used for
    forestry monitoring to read sensors in the forest canopy.  }
\label{fig:tree-read}
\end{figure}

In Figure \ref{fig:tree-read}, we show a RC-controlled drone reading a light-sensing tag affixed to a tree. As the price of sensorized tags approaches that of their ID-only counterparts (\$0.10 ea, or lower), we may start to see situations in which monitoring crops on a plant-by-plant basis becomes viable.  Such sensors could measure plant health, incident solar radiation, water levels, fruit ripeness, etc.  Indeed, this is an exciting area to pursue as sensor tags become pervasive.

\section{ Discussion \& Conclusions }

In this paper, we provided an overview of recently-commercialized sensor tags, and for the first time, we showed how robots might utilize such sensor tags. We demonstrated a prototype application for ``smart field'' soil moisture monitoring with some basic experimental evaluation. Finally, we demonstrated some basic setups that demonstrate compelling future possibilities of sensorized tags.

While the possibilities are compelling, current long-range RFID systems are not a panacea. Perpetual concerns like read range, power budgets, and robot control still play a significant role.  Even if ameliorated by battery-assist or energy-harvesting technologies, there will still be significant challenges associated with building cost-effective sensorized tags and integrating them directly into objects and the environment.

\section{ Acknowledgments }

We would like to thank Matt Reynolds, John O'Hollaren, and Sid Kandan for their insightful discussions on this topic.

\bibliographystyle{IEEEtran}
\bibliography{quad,tdeyle,hrl_all,rfid}

\end{document}